\documentclass[letterpaper, 10 pt, conference]{ieeeconf}  

\usepackage{amsmath}
\usepackage{amsfonts}
\usepackage{amssymb}
\usepackage{algorithm}
\usepackage[noend]{algpseudocode}
\usepackage{xcolor}
\usepackage{graphicx}
\usepackage{subcaption}
\usepackage{booktabs}
\usepackage[capitalize]{cleveref}
\usepackage{multirow}

\IEEEoverridecommandlockouts                              

\title{\LARGE \bf
Sample-Efficient Expert Query Control in Active Imitation Learning \\ via Conformal Prediction
}

\author{Arad Firouzkouhi$^{1}$, Omid Mirzaeedodangeh$^{2}$, and Lars Lindemann$^{2}$
\thanks{$^{1}$ Department of Computer Science, University of Southern California, Los Angeles, CA 90089, USA
        {\tt\small firouzko@usc.edu}}%
\thanks{$^{2}$ Department of Information Technology and Electrical Engineering, ETH Zurich, 8092 Zurich, Switzerland
        {\tt\small \{omirzaeedoda, llindemann\}@ethz.ch}}%
} 

\begin{document}

\maketitle
\thispagestyle{empty}
\pagestyle{empty}

\begin{abstract}

Active imitation learning (AIL) combats covariate shift by querying an expert during training. However, expert action labeling often dominates the cost, especially in GPU-intensive simulators, human-in-the-loop settings, and robot fleets that revisit near-duplicate states. We present \emph{\mbox{Conformalized Rejection Sampling} for Active Imitation Learning (CRSAIL)}, a querying rule that requests an expert action only when the visited state is under-represented in the expert-labeled dataset. 
CRSAIL scores state novelty by the distance to the \mbox{\(K\)-th} nearest expert state and sets a single global threshold via conformal prediction. This threshold is the empirical \((1-\alpha)\) quantile of on-policy calibration scores, providing a distribution-free calibration rule that links \(\alpha\) to the expected query rate and makes \(\alpha\) a task-agnostic tuning knob.
This state-space querying strategy is robust to outliers and, unlike safety-gate-based AIL, can be run without real-time expert takeovers: we roll out full trajectories (episodes) with the learner and only afterward query the expert on a subset of visited states. Evaluated on MuJoCo robotics tasks, CRSAIL matches or exceeds expert-level reward while reducing total expert queries by up to 96\% vs.\ DAgger and up to 65\% vs.\ prior AIL methods, with empirical robustness to \(\alpha\) and \(K\), easing deployment on novel systems with unknown dynamics.

\end{abstract}

\section{Introduction}
Imitation learning (IL) offers a compelling alternative to reward engineering in reinforcement learning by training a policy to reproduce expert behavior directly from demonstrations~\cite{osa2018algorithmic}. Its successes range from dexterous robot control to autonomous driving~\cite{hu2022model,ross2013learning,kim2024surgical}. However, IL suffers from \emph{covariate shift}: as the agent explores, its on-policy state distribution diverges from the expert’s, leading to compounding control errors~\cite{ross2011reductionimitationlearningstructured}. \emph{Active imitation learning} (AIL) addresses this by querying the expert for additional action labels in states where the learner is likely to fail. In practice, those queries can be the chief bottleneck: running a high-fidelity simulator for an expert consumes GPU hours, human-in-the-loop labeling induces operator fatigue, and safety-critical domains may forbid frequent interventions~\cite{laskey2017comparing,zhang2016query}.

Most existing query-management schemes either ask the expert far too often~\cite{bcfail}, require the expert to seize full control~\cite{hoque2021lazydaggerreducingcontextswitching,hoque2021thriftydaggerbudgetawarenoveltyrisk}, or rely on action-uncertainty thresholds~\cite{menda2019ensembledaggerbayesianapproachsafe} that do not reliably indicate novelty in state space. Consequently, they inflate both the annotation budget and the aggregated expert dataset with redundant data, slowing training while adding little informational value. For example, multi-agent systems executing the same policy do not need to query an expert again if a similar state has already been labeled; a naive approach would increase the number of expert queries roughly in proportion to the number of agents~\cite{hoque2023fleet}.
 \begin{figure}[t!]
    \centering    
    \includegraphics[width=\linewidth]{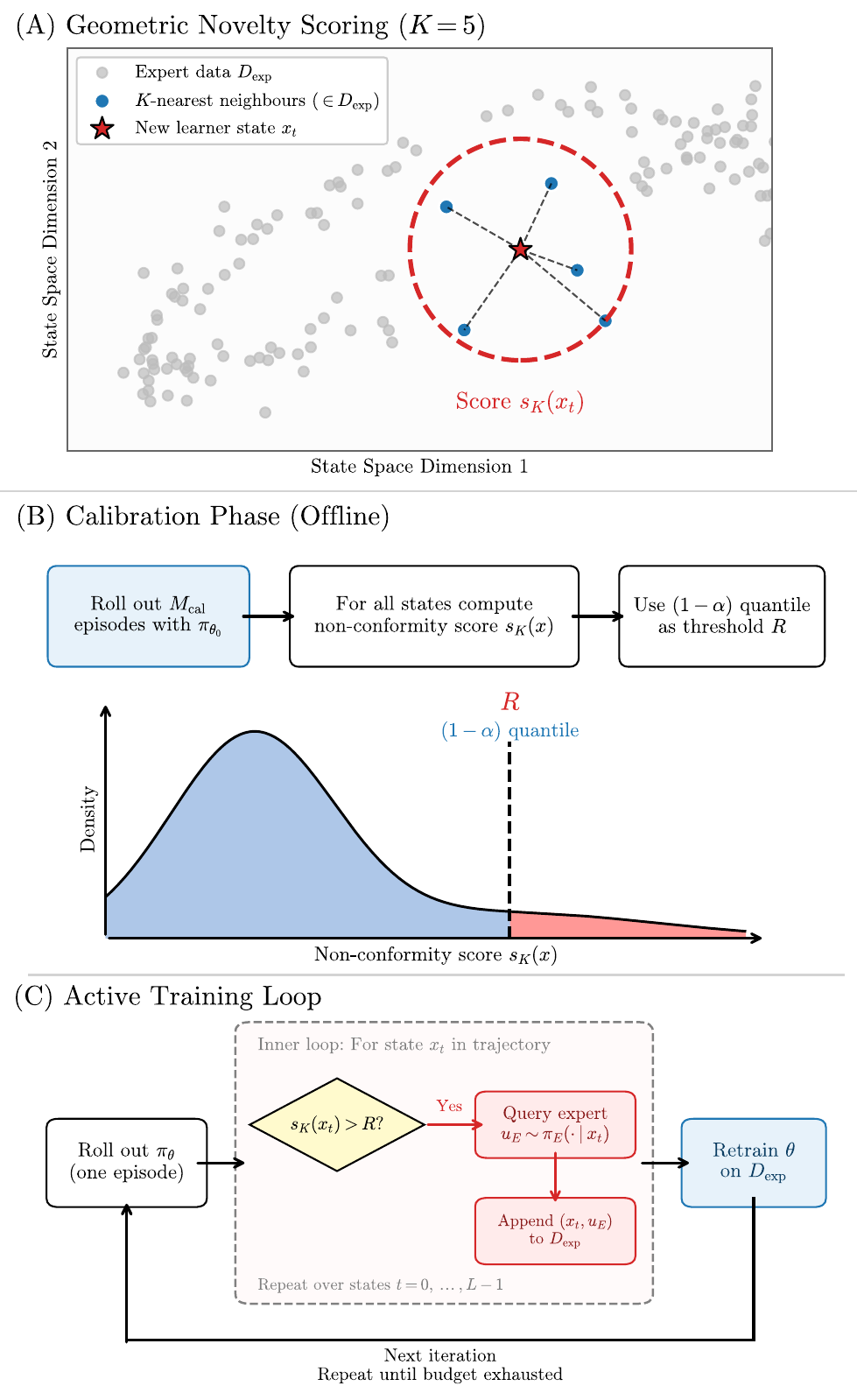}
    \caption{
Overview of CRSAIL.
(A) We assign each learner state a geometric novelty score \(s_K(x_t)\) based on the distance to the
\(K\)-nearest neighbours in the expert dataset \(D_{\mathrm{exp}}\).
(B) Offline, we roll out an initial behavior cloning policy, compute scores for visited states, and
set a single query threshold \(R\) as the \((1-\alpha)\)-quantile of this
distribution.
(C) Online, during training, we roll out our learner policy for one trajectory, and query the expert where \(s_K(x_t) > R\).
We add the labeled state to \(D_{\mathrm{exp}}\), and retrain the
policy.
}
\label{fig:placeholder}
\end{figure}
We address this challenge with \textbf{Conformalized Rejection Sampling for Active Imitation Learning} (CRSAIL). CRSAIL casts selective querying as a conformal prediction problem and uses conformal calibration to set a data-driven query threshold. To quantify how new an encountered state is, we use the distance to the \(K\)-th nearest neighbor in the existing expert-labeled set: a large distance indicates that the agent has entered an unfamiliar region of the state space. During an initial calibration phase, we roll out the initial policy to collect unlabeled on-policy states and set a distance threshold \(R\) as an empirical \((1-\alpha)\) quantile of their nonconformity scores (see \cref{fig:placeholder} for an overview). This threshold turns \(\alpha\) into a principled tuning knob for controlling the expected query rate. At training time, if a visited state’s distance lies below \(R\) (a region already well represented by expert data), we forgo querying; otherwise, we request the expert action. Queries are issued post hoc in batch after each episode, so no real-time takeovers are required. This simple rule concentrates expert effort on truly novel and poorly covered regions. As a result, it reduces labeling cost and promotes a more diverse expert dataset by populating the finite replay buffer with a compact set of informative states rather than near-duplicates. Conformal prediction also makes threshold selection robust; because \(R\) is defined by a quantile, it is not skewed by extreme outliers caused by covariate shift. By choosing a miscoverage rate \(\alpha\), we obtain principled budget control: the fraction of queried states is approximately \(\alpha\). This data-driven threshold adapts to the state-space density of each task and reduces per-task retuning.

The remainder of this paper is organized as follows.
We will first introduce existing methods that reduce expert queries, 
then formulate density-based sample rejection as a conformal prediction problem, detail the CRSAIL procedure, present experiments and ablations, and conclude with limitations and future directions.

\section{Related Work}
\textbf{\emph{Behavioral cloning}} treats imitation learning as a supervised learning problem on an offline expert dataset~\cite{bain1995framework}. Under covariate shift, small errors push the policy into unseen states where mistakes grow and compound~\cite{bcfail}.

\textbf{\emph{Interactive imitation learning}} queries the expert for corrective action during training to compensate for covariate shift. Dataset Aggregation (DAgger) rolls out the current policy, labels \emph{every} visited state with the expert action, aggregates the new labels with prior data, and retrains~\cite{ross2011reductionimitationlearningstructured}. This label-all strategy is effective but \emph{query-inefficient}.

\textbf{\emph{Active imitation learning}} uses \emph{gating} to control annotation costs by requesting expert labels for only a subset of visited states. The goal is to focus supervision on states that are informative for improving the learner while avoiding redundant labels in well-covered regions of the state space. An early approach was SafeDAgger~\cite{zhang2016query} which learns an auxiliary safety classifier that predicts whether a state is safe enough for the learner to act upon. If not, the expert will take over until it is safe again, and these expert-controlled states will be added to the dataset. SafeDAgger’s gains come from labeling only predicted-unsafe states; however, the safety policy itself must be maintained and tuned, and its conservatism can still induce frequent interventions. 

\textbf{\emph{Uncertainty and safety-gate-based algorithms}} like SafeDAgger~\cite{zhang2016query} learn a safety classifier that decides whether the learner may act; otherwise the expert takes control and those states are labeled. Moreover, LazyDAgger~\cite{hoque2021lazydaggerreducingcontextswitching} queries based on a secondary network’s estimate of action uncertainty. Furthermore, EnsembleDAgger~\cite{menda2019ensembledaggerbayesianapproachsafe} and ThriftyDAgger~\cite{hoque2021thriftydaggerbudgetawarenoveltyrisk} use disagreement or variance across multiple policies, with fixed and percentile-based thresholds respectively. ThriftyDAgger also added a Q-network for safety assessment. Action-space disagreement mixes epistemic uncertainty, which active methods aim to reduce with data, with aleatoric randomness from stochastic dynamics or multiple valid actions; the latter can be high even in well-covered regions, which weakens action-based novelty tests and can lead to unnecessary queries. CRSAIL aims to directly measure state-space novelty instead.

\textbf{\emph{Algorithms based on Random Network Distillation}} like RNDAgger~\cite{bire2025efficientactiveimitationlearning} uses the prediction error of a randomly initialized target network as a state-space novelty score and query whenever this error exceeds a threshold. This requires training auxiliary networks and tuning a per-task threshold, and the resulting prediction errors can be sensitive to stochasticity in the observations, causing repeated queries in well-covered regions of the state space. By measuring state space coverage directly via geometry and using a single conformally calibrated threshold, CRSAIL runs rollouts end-to-end with the learner and queries the expert post hoc in batch after each episode with no real-time takeover needed.  This is in direct contrast with all mentioned methods except EnsembleDAgger.

\textbf{\emph{Conformal methods in imitation learning}} have recently been used in ConformalDAgger~\cite{zhao2025conformalizedinteractiveimitationlearning}, but for a fundamentally different purpose. Its goal is not query efficiency but robustness to shifts in the expert's policy over time. It calibrates uncertainty over the expert's actions to detect expert policy drift. In contrast, our method uses conformal prediction to assess novelty in the state space, allowing the agent to avoid redundant queries in regions already well represented by existing data.
The two approaches are complementary and address different challenges.
\section{Problem Formulation}
\label{sec:problem}
\paragraph{Environment}
We model the environment as a discrete-time Markov decision process
\begin{equation}
\label{eq:mdp}
M \;:=\; \big(\mathcal{X},\, \mathcal{U},\, P,\, r,\, \mathcal{X}_0,\, \mathcal{X}_T,\, T_{\max}\big),
\end{equation}
where
\(\mathcal{X}\) is a measurable state space,
\(\mathcal{U}\) is an action space,
\(P(\cdot \mid x,u)\) is a Markov transition kernel on \(\mathcal{X}\),
\(r:\mathcal{X}\times\mathcal{U}\to\mathbb{R}\) is a reward used only for evaluation,
\(\mathcal{X}_0\) is a distribution on \(\mathcal{X}\) for initial states,
\(\mathcal{X}_T\subseteq\mathcal{X}\) is a terminal set,
and \(T_{\max}\in\mathbb{N}\cup\{\infty\}\) is the episode horizon.
A policy may be deterministic \(\pi:\mathcal{X}\to\mathcal{U}\) or stochastic, with \(\pi(\cdot\mid x)\) being a distribution with support over \(\mathcal{U}\).
An episode begins at \(x_0\sim \mathcal{X}_0\) and evolves, for \(t=0,1,\ldots\), as

\begin{align}
u_t &\sim \pi(\cdot \mid x_t),\quad
x_{t+1} \sim P(\cdot \mid x_t, u_t). \label{eq:dynamics}
\end{align}
The random episode length is
\begin{equation}
\label{eq:episode_length}
L \;:=\; \min\big\{\, t\ge 1 \,:\, x_t\in \mathcal{X}_T \ \text{or}\ t = T_{\max} \,\big\}.
\end{equation}
Rolling out a policy $\pi$ under the dynamics in~\cref{eq:dynamics} yields the random trajectory
\begin{equation}
\label{eq:trajectory}
  \tau_\pi := (x_0, u_0, \ldots, x_{L-1}, u_{L-1}, x_L).
\end{equation}
\paragraph{Expert and imitation loss}
We are given an expert policy \(\pi_E\), possibly stochastic. When the expert is queried at state \(x\), the expert action label is a draw \(u_E\sim \pi_E(\cdot\mid x)\).
Let \(\pi_\theta\) be a parametric learner and \(\ell:\mathcal{U}\times\mathcal{U}\to\mathbb{R}_{\ge 0}\) an action loss (e.g., squared error).
We define the population imitation loss as
\begin{align}
\label{eq:imitation_loss}
J(\theta)
&:=\; \mathbb{E}\!\left[ \sum_{t=0}^{L-1} \ell\big(\pi_\theta(x_t),\, u_{E,t}\big) \right],
\end{align}
where the expectation is with respect to \(x_0\sim\mathcal{X}_0\), the trajectory generated by \(P\) and \(\pi_\theta\), and the expert draws \(u_{E,t}\sim\pi_E(\cdot\mid x_t)\).
An extension to stochastic learner policies $\pi_\theta(\cdot\mid x)$ is possible
with minor changes of notation; $\pi_\theta$ is deterministic in our experiments.
The expert labels \(u_{E,t}\) are a conceptual oracle for defining the objective; during training we only observe labels at queried times (defined below).
Let \(\theta^\star \in \arg\min_\theta J(\theta)\) denote an ideal minimizer; computing \(\theta^\star\) is generally infeasible due to unknown dynamics and nonconvexity.
Our algorithms therefore learn an approximate solution from a subset of expert action labels.

\paragraph{Initial dataset}
For a given target of \(M\) expert state–action pairs, the initial expert dataset concatenates \(n_M\) expert rollouts \(\{\tau_{\pi_E}^{(e)}\}_{e=1}^{n_M}\) whose lengths \(\{L_{e}\}\) satisfy \(\sum_{e=1}^{n_M} L_{e} \ge M\), where each $\tau_{\pi_E}^{(e)}$ is a trajectory
of $\pi_E$ as in~\eqref{eq:trajectory}, and datasets are viewed as multisets of
state–action pairs with $\uplus$ denoting multiset union. We define
\begin{equation}
\label{eq:initial_dataset}
D_{\mathrm{exp}}^{(0)}
  := \biguplus_{e=1}^{n_M}
     \big\{ (x^{(e)}_t, u^{(e)}_t) : t = 0,\ldots,L_e - 1 \big\}.
\end{equation}
We obtain our initial learner parameter $\theta_0$ and policy $\pi_{\theta_0}$ via
behavioral cloning (BC) on \(D_{\mathrm{exp}}^{(0)}\).

\paragraph{Admissible querying strategies}
Training proceeds in episodes \(i=0,1,2,\ldots\).
At the start of episode \(i\), the dataset is \(D_{\mathrm{exp}}^{(i)}\) and the learner is \(\pi_{\theta_i}\).
We roll out \(\pi_{\theta_i}\) to obtain \(\tau^{(i)}=(x^{(i)}_0, u^{(i)}_0, \ldots, x^{(i)}_{L_i})\).
Let the training history before episode \(i\) be
\begin{equation}
\label{eq:history}
H_i :=
\big( D_{\mathrm{exp}}^{(0)}, \tau^{(0)}, S_0, Q_0, \ldots, \tau^{(i-1)}, S_{i-1}, Q_{i-1} \big)
\end{equation}
where \(S_j\) denotes the query index set and \(Q_j\) the corresponding
expert-labeled state–action pairs for episode \(j\), as defined next.
First, a \emph{querying strategy} \(\psi\) specifies, for each episode, at which
time steps the expert is queried. Given the training history \(H_i\) in
\eqref{eq:history} and the current trajectory
\(\tau^{(i)}=(x^{(i)}_0,u^{(i)}_0,\ldots,x^{(i)}_{L_i})\), we define the
\emph{query index set} for episode \(i\) as
\begin{equation}
\label{eq:psi}
S_i \;:=\; \psi\big(H_i, \tau^{(i)}\big) \;\subseteq\; \{0,\ldots,L_i-1\},
\end{equation}
so that \(t\in S_i\) means that the state \(x^{(i)}_t\) is sent to the expert
for labeling.
We require \(\psi\) to be non-anticipatory across episodes (it does not depend
on future episodes).
Given \(S_i\), the per episode query multiset is
\begin{equation}
\begin{aligned}
\label{eq:Qi}
Q_i
&:= \big\{
\big(x^{(i)}_t, u^{(i)}_{E,t}\big):\,
t\in S_i,\;
u^{(i)}_{E,t} \sim \pi_E(\cdot\mid x^{(i)}_t)
\big\}.
\end{aligned}
\end{equation}
For reference, DAgger corresponds to \(S_i=\{0,\ldots,L_i-1\}\) for all \(i\).

\paragraph{Dataset update and learner update}
After episode \(i\), we update the dataset by multiset union and update the learner by a learning operator:
\begin{equation}  
\label{eq:dataset_update}
\begin{aligned}
D_{\mathrm{exp}}^{(i+1)}
:=\; D_{\mathrm{exp}}^{(i)} \ \uplus\ Q_i\;,\;
\theta_{i+1} \gets \mathsf{Update}\big(\theta_i, D_{\mathrm{exp}}^{(i+1)}\big)
\end{aligned}
\end{equation}
where \(\mathsf{Update}\) is a generic parameter–update operator (e.g., multiple gradient descent steps) on the empirical imitation loss  (of \cref{eq:imitation_loss}) over \(D_{\mathrm{exp}}^{(i+1)}\).
This produces sequences \(\{\theta_i\}_{i\ge 0}\) and \(\{\pi_{\theta_i}\}_{i\ge 0}\).

\paragraph{Stopping time and training length \(C\)}

For a fixed querying strategy \(\psi \in \Psi\) and budgets
\(B\) (queries) and \(T_{\mathrm{train}}\) (steps) in
\(\mathbb{N}\cup\{\infty\}\),
we define the number of training episodes (which is a random variable) as
\begin{equation}
\label{eq:C_def}
C \;:=\; \min \Big\{\, i \ge 1 :
\sum_{j=0}^{i-1} |Q_j| \,\ge\, B
\ \text{or}\
\sum_{j=0}^{i-1} L_j \,\ge\, T_{\mathrm{train}} \Big\}.
\end{equation}
Training halts once either the total number of queries reaches \(B\)
or the total number of environment steps reaches \(T_{\mathrm{train}}\);
setting \(B=\infty\) or \(T_{\mathrm{train}}=\infty\) relaxes
the corresponding budget constraint.
By construction, \(C\) is a stopping time with respect to the natural
training history, as it depends only on information available up to episode \(i\).
\paragraph{Objective and the optimal strategy \(\psi^\star\)}
Let \(\Psi\) denote the class of admissible strategies, as defined per \cref{eq:psi}.
Our goal is to minimize the expected imitation loss of the final policy under budgets \(B\) and \(T_{train}\).
We therefore aim to solve
\begin{align}
\label{eq:opt_expected_budget}
\psi^\star \;\in\; \arg\min_{\psi\in\Psi}\ \ &
\mathbb{E}\!\left[ J\big(\theta_{C}\big) \right],
\end{align}
where the expectation is over the joint randomness of \(\mathcal{X}_0\), \(P\), the learner policies \(\{\pi_{\theta_i}\}\), the expert \(\pi_E\), the strategy \(\psi\), and the learning operator \(\mathsf{Update}\).
Both the stopping time \(C\) and the terminal parameters \(\theta_C\) depend on the chosen strategy \(\psi\); we suppress this dependence in the notation for
readability.

\paragraph{What \(C\), \(Q\), and \(\psi^\star\) mean operationally}
The stopping time \(C\) is the data dependent training length induced by \eqref{eq:C_def}; it is the first episode at which either the query budget or
the interaction budget is exhausted.
The multiset \(Q\) is the realized collection of expert action labels; its cardinality \(\sum_{i=0}^{C-1}|Q_i|\) is the consumed expert budget.
The strategy \(\psi^\star\) is the rule that,
for given budgets \((B, T_{\mathrm{train}})\), yields the lowest expected final
imitation loss among all admissible strategies. One may either fix a query budget \(B\) (and take
\(T_{\mathrm{train}}=\infty\)) and compare the achieved reward at that budget,
or fix an interaction budget \(T_{\mathrm{train}}\) (and take \(B=\infty\)) and
compare the total number of queries and the reward. The former is more natural in deployment, while the latter is convenient for analysis and is the regime we use in
Sec.~\ref{sec:experiments}.

\paragraph{Intuition for the strategy class used later}
In Section~\ref{sec:method}, we introduce episode-level querying strategies that use state-space coverage to decide when to query the expert. For integers \(K\ge 1\) and a threshold \(R>0\), define
\begin{equation}
\begin{aligned}
\label{eq:psi_RK}
\psi_{R,K}(H_i,\tau^{(i)})
&:=\ \big\{\, t\in\{0,\ldots,L_i-1\} \\&\,:\ 
s_K\!\big(x^{(i)}_t;\, D^{(i)}_{\mathrm{exp}}\big) > R \,\big\},
\end{aligned}
\end{equation}
where the novelty score \(s_K(x; D_{\mathrm{exp}}^{(i)})\) is the distance from \(x\) to its \(K\)-th nearest neighbor among the states in the current expert dataset \(D_{\mathrm{exp}}^{(i)}\). We set \(R\) once by conformal calibration on unlabeled on-policy states so that a user-specified miscoverage \(\alpha\) directly controls the expected query rate.
Because \(\psi_{R,K}\) operates post hoc at the episode level, it requires no real-time expert takeovers and belongs to the admissible class \(\Psi\).


\section{Conformalized Rejection Sampling for Active Imitation Learning}
\label{sec:method}
To solve the optimization problem in \cref{eq:opt_expected_budget}, we instantiate an admissible post hoc querying strategy \(\psi_{R,K}\) (see \cref{eq:psi_RK}) that avoids requesting expert action labels in regions of the state space already well represented by the current expert dataset \(D_{\mathrm{exp}}^{(i)}\). The strategy is evaluated after each episode, produces the per-episode query multisets \(Q_i\) in the form of \cref{eq:Qi}, and, together with the learning operator \(\mathsf{Update}\), determines the stopping time \(C\) in \cref{eq:C_def}.

\subsection{Querying by state-space novelty}
\label{subsec:novelty}
Let the state projection of the current expert dataset be
\begin{equation}
\label{eq:DX_def}
D_X^{(i)} \;:=\; \big\{\, x' : (x',u') \in D_{\mathrm{exp}}^{(i)} \,\big\}.
\end{equation}
For a norm \(\|\cdot\|\) on \(\mathbb{R}^d\), define the nonconformity score as the distance to the \(K\)-th nearest expert state:
\begin{equation}
\label{eq:sK_def}
s_K\big(x;D_{\mathrm{exp}}^{(i)}\big)
:= \inf\{\, r\!\ge\!0 : \lvert B(x,r)\cap D_X^{(i)}\rvert \ge K \},
\end{equation}
where \(B(x,r)=\{z\in\mathbb{R}^d:\|z-x\|\le r\}\).
In the language of conformal prediction, \(s_K\) is a \emph{nonconformity
score}: larger values mean that \(x\) is more atypical relative to the expert
data.
Equivalently, \(s_K\big(x;D_{\mathrm{exp}}^{(i)}\big)\) is the radius of the smallest closed ball centered at \(x\) that contains at least \(K\) elements of \(D_X^{(i)}\). Larger scores indicate lower local data density and hence greater novelty; increasing \(K\) makes the score more robust to outliers and accepting of duplicates.

Given a threshold \(R>0\), the post hoc query indices and the queried multiset for episode \(i\) are
\begin{equation}
\label{eq:psi_rk_inst}
 \begin{aligned}
S_i
&=\; \psi_{R,K}\!\big(H_i,\tau^{(i)}\big)
\\&:= \big\{\, t \in \{0,\ldots,L_i-1\} : s_K\!\big(x_t^{(i)}; D_{\mathrm{exp}}^{(i)}\big) > R \,\big\},\\
Q_i
&=\; \big\{\, (x_t^{(i)}, u^{E,(i)}_t) : t \in S_i \big\},
\end{aligned}   
\end{equation}
which match the general definitions in \cref{eq:psi,eq:Qi}.

\paragraph*{Intuition}
The rule in \cref{eq:psi_rk_inst} concentrates expert effort on under-covered regions: if many expert states already lie in a small neighborhood of \(x_t^{(i)}\), we skip labeling; if the neighborhood is sparse, we request the expert label. This targets the query budget at states expected to improve generalization while avoiding redundant labels in well-represented areas.
\subsection{Threshold selection via conformal calibration}
\label{subsec:conformal}
The key question is \emph{“How do we set a single threshold \(R\) in a task-agnostic and statistically principled way?"} We calibrate \(R\) using conformal prediction on unlabeled \emph{on-policy} states from the initial learner \(\pi_{\theta_0}\) (no expert labels are needed for calibration). We briefly recall only the ingredients needed here and refer the reader to
standard introductions to conformal prediction for a more comprehensive
treatment; see \cite{angelopoulos2022gentleintroductionconformalprediction,lindemann2024formal}.
\paragraph*{Calibration dataset}

We roll out $\pi_{\theta_0}$ for $M_{\mathrm{cal}}$ episodes to obtain
trajectories $\{\tau_{\pi_{\theta_0}}^{(e)}\}_{e=1}^{M_{\mathrm{cal}}}$ as in
\eqref{eq:trajectory}, with lengths $\{L_e\}$. Define the calibration multiset
of visited on-policy states as
\begin{equation}
\label{eq:Xcal_def}
X_{\mathrm{cal}}
:= \biguplus_{e=1}^{M_{\mathrm{cal}}}
\{ x^{(e)}_t : t = 0,\ldots,L_e - 1 \}
= \{ x_j \}_{j=1}^{N_{\mathrm{cal}}},
\end{equation}
where $N_{\mathrm{cal}} = \sum_{e=1}^{M_{\mathrm{cal}}} L_e$ is the total
number of calibration states.

For each \(x_j\in X_{\mathrm{cal}}\) compute
\begin{equation}
\label{eq:scores_cal}
s_j \;:=\; s_K\!\big(x_j;\, D_{\mathrm{exp}}^{(0)}\big),
\qquad j=1,\ldots,N_{\mathrm{cal}}.
\end{equation}

\paragraph*{Finite-sample quantile}
For a user-chosen miscoverage \(\alpha\in(0,1)\), define
\begin{equation}
\label{eq:m_index}
m \;:=\; \left\lceil (N_{\mathrm{cal}}+1)\,(1-\alpha)\right\rceil,
\end{equation}
and let \(s_{(1)} \le \cdots \le s_{(N_{\mathrm{cal}})}\) be the order statistics of \(\{s_j\}\).
Set the calibrated threshold
\begin{equation}
\label{eq:R_def}
R \;:=\; s_{(m)} .
\end{equation}
In practice, \cref{eq:R_def} is equivalent to a non-interpolating empirical quantile of level \(q=m/N_{\mathrm{cal}}\).

\paragraph*{Coverage guarantee and interpretation}
If the calibration states and future on-policy states were exchangeable for a
fixed policy, classical conformal prediction would imply
\begin{equation}
\label{eq:coverage}
\Pr\!\big[\, s_K(x_{\mathrm{new}};\,D_{\mathrm{exp}}^{(0)}) \le R \,\big]
\;\ge\; 1-\alpha,
\end{equation}
for any future on-policy state \(x_{\mathrm{new}}\). Thus, a fraction of
at least \(1-\alpha\) of such states would lie in \(R\)-dense regions and would
not be queried by \cref{eq:psi_rk_inst}, so \(\alpha\) specifies a nominal
query rate. In our actual training procedure the policy evolves over
iterations and, even for a fixed policy, states along a trajectory are
temporally correlated and not identically distributed across time steps, so
exchangeability is violated and \eqref{eq:coverage} should be viewed as an
idealized reference rather than a strict guarantee. Nevertheless, varying
\(\alpha\) affects the empirical quantile \(R\): increasing
\(\alpha\) lowers the target coverage level, thus decreases \(R\) and marks a
larger fraction of states as “novel,” leading to more expert queries. \emph{This
makes \(\alpha\) an effective knob for trading off coverage and query rate} even
without a formal guarantee under policy shift.

\paragraph*{Why not recalibrate with an updated policy?}

As learning progresses, the learner policy \(\pi_{\theta_i}\) typically spends
more time in regions of the state space that are already well covered by the
aggregated expert dataset \(D_{\mathrm{exp}}^{(i)}\). One might consider
periodically re-running the conformal calibration step with the current policy
and dataset to obtain updated thresholds \(R_i\). However, this continually
re-normalizes the distance scores to the states that are currently visited, forcing the query rate to remain approximately constant rather than decaying as coverage improves.
Such periodic recalibration largely removed the desirable decay of the query
rate over training duration. For
this reason we calibrate once using \(\pi_{\theta_0}\) and keep a single
threshold \(R\) fixed; as \(\pi_{\theta_i}\) improves and \(D_{\mathrm{exp}}^{(i)}\)
grows around expert-like regions, an increasing fraction of visited states
falls within the region where \(s_K(x; D_{\mathrm{exp}}^{(i)}) \le R\) and no
longer triggers queries.

\begin{algorithm}[ht]
    \caption{CRSAIL radius calibration}
    \label{alg:crsail_calibration}
    \begin{algorithmic}[1]
        \State \textbf{Inputs:} initial learner \(\pi_{\theta_0}\); expert dataset \(D_{\mathrm{exp}}^{(0)}\); integer \(K\); miscoverage \(\alpha\in(0,1)\); calibration episodes \(M_{\mathrm{cal}}\).
        \State Roll out \(\pi_{\theta_0}\) for \(M_{\mathrm{cal}}\) episodes; collect \(X_{\mathrm{cal}}=\{x_j\}_{j=1}^{N_{\mathrm{cal}}}\).
        \For{$j=1$ to $N_{\mathrm{cal}}$}
            \State $s_j \gets s_K\big(x_j;\, D_{\mathrm{exp}}^{(0)}\big)$ \Comment{distance to the \(K\)th neighbor}
        \EndFor
        \State $m \gets \left\lceil (N_{\mathrm{cal}}+1)(1-\alpha) \right\rceil$
        \State $R \gets \textsf{OrderStatistic}\big(\{s_j\},\, m\big)$
        \State \textbf{return} $R$
    \end{algorithmic}
\end{algorithm}

\subsection{The CRSAIL training algorithm}
\label{subsec:crsail_algo}
After behavioral cloning on \(D_{\mathrm{exp}}^{(0)}\) and a single calibration to compute \(R\), CRSAIL iterates episodes with post hoc batch querying and dataset aggregation. Algorithms~\ref{alg:crsail_calibration} and
\ref{alg:crsail} summarize the radius-calibration step and the full CRSAIL
training loop, respectively. Each iteration instantiates \(\psi_{R,K}\) and forms \(Q_i\) via \cref{eq:psi_rk_inst}, updates \(D_{\mathrm{exp}}^{(i+1)}\) and advances the learner by \cref{eq:dataset_update}. Training halts at the stopping time \(C\) in \eqref{eq:C_def}, i.e., as soon as
either the query budget \(B\) or the step budget \(T_{\mathrm{train}}\) is
exhausted.
\begin{algorithm}[ht]
\caption{CRSAIL}
\label{alg:crsail}
\begin{algorithmic}[1]
\State \textbf{Inputs:} expert policy \(\pi_E\); initial dataset \(D_{\mathrm{exp}}^{(0)}\);
miscoverage \(\alpha\); integer \(K\) \textit{(K-th neighbor order)};
calibration episodes \(M_{\mathrm{cal}}\);
query budget \(B\); step budget \(T_{\mathrm{train}}\).
\State \(\theta_0 \gets \textsf{BehavioralCloning}\big(D_{\mathrm{exp}}^{(0)}\big)\)
\State \(R \gets \textsf{CRSAIL\_Calibration}\big(\pi_{\theta_0}, D_{\mathrm{exp}}^{(0)}, K, \alpha, M_{\mathrm{cal}}\big)\)
\State \(i \gets 0,\quad t \gets 0,\quad q \gets 0\) \Comment{iterations, steps and queries}
\While{\(t < T_{\mathrm{train}}\) \textbf{and} \(q < B\)}
  \State Roll out \(\pi_{\theta_i}\) to obtain states \(x^{(i)}_0,\ldots,x^{(i)}_{L_i}\)
  \State \(S_i \gets \big\{\, t \in \{0,\ldots,L_i-1\} : s_K(x^{(i)}_t; D_{\mathrm{exp}}^{(i)}) > R \,\big\}\)
  \State \(Q_i \gets \big\{\, \big(x^{(i)}_t, u^{(i)}_{E,t}\big) : t\in S_i \,\big\}\)
  \State \(D_{\mathrm{exp}}^{(i+1)} \gets D_{\mathrm{exp}}^{(i)} \,\uplus\, Q_i\)
  \State \(\theta_{i+1} \gets \textsf{UPDATE}\big(\theta_i, D_{\mathrm{exp}}^{(i+1)}\big)\)
  \State \(t \gets t + L_i\)
  \State \(q \gets q + |Q_i|\)
  \State \(i \gets i+1\)
\EndWhile
\State \textbf{return} \(\pi_{\theta_i}\) \Comment{final policy \(\pi_{\theta_C}\)}
\end{algorithmic}
\end{algorithm}
\paragraph*{Computational notes}
We construct tensors for the learner states and expert states and use the GPU to compute their pairwise distance matrix. We then take per-state K-th-order statistics. Since K is small and the queries are post hoc, this adds only minor overhead compared with simulation.
\paragraph*{Link back to the objective}
CRSAIL sets \(\psi=\psi_{R,K}\) in \cref{eq:opt_expected_budget}. In our
experiments, when $\Psi$ is restricted to CRSAIL and the baseline strategies,
$\psi_{R,K}$ consistently achieves near-expert reward while using dramatically
fewer expert queries, so it
serves as an empirical surrogate for $\psi^\star$ within this restricted class.

\section{Experiments}
\label{sec:experiments}
\subsection{Experimental Setup}
We evaluate on three MuJoCo control tasks for stabilization, manipulation, and locomotion  with randomized initial states. The expert for each task is an RL model trained with Stable Baselines3 to near-convergence. We compare \textbf{CRSAIL} against \textbf{DAgger}, \textbf{EnsembleDAgger}, and \textbf{ThriftyDAgger}. We omit SafeDAgger, LazyDAgger, and RNDAgger: SafeDAgger’s gating is subsumed by later baselines, while LazyDAgger (by the same authors as Thrifty) was highly sensitive and unstable in our pilots (often failing to query and stagnating). Thrifty demonstrably improves upon LazyDAgger, so we treat Thrifty as its successor baseline. RNDAgger had no public implementation, and any speculative reimplementation would risk an unfair comparison.

All learned policies (and auxiliary networks) use an MLP with a single hidden
layer of size $64$. After each training episode, we evaluate the learner over
100 episodes and compute its average episodic return; a run is deemed to have
\emph{converged to expert level} once this average reaches at least $95\%$ of
the expert’s. For each baseline, we use the hyperparameters reported in the
original paper on any environment they benchmarked; on environments not covered
there, we sweep the gating hyperparameters to span query rates from near $0\%$
upward and choose the smallest value that yields convergence on most runs (for
a fixed offline dataset), in order to maximize query efficiency. The final
choices are listed in \cref{tab:hyperparams_env_keyval}. For CRSAIL we fix $K=5$ and choose $\alpha$ from a coarse sweep, selecting a
slightly conservative value in the range $\alpha \in [0.9, 0.95]$ where
performance and query counts are stable (see \cref{tab:alpha_sweep}).
We report (i) convergence rate, (ii) the number of expert queries issued until
the run first reaches expert-level performance, and (iii) the total number of
expert queries over the entire training run. For each dataset size $M$, we create five independent offline datasets drawn from the same expert as discussed in \cref{eq:initial_dataset}.
All methods are run with a fixed interaction budget \(T_{\mathrm{train}}\) and \(B=\infty\). Thus all methods see
the same number of environment steps, while their total expert queries
\(\sum_{i=0}^{C-1} |Q_i|\) may differ.

\emph{Task 1: Inverted Double Pendulum (InvDP)}. We apply horizontal force to balance a double pendulum (9-dimensional state space,  scalar action space). Episodes last up to 1{,}000 steps with early termination on failure, emphasizing  query efficiency when trajectories bifurcate to quick failures with zero reward vs. long sequences with large rewards. This causes  challenges as our expert is imperfect ($90\%$ success rate). Offline datasets use $M\!\in\!\{1\text{k},2\text{k},3\text{k},5\text{k},10\text{k}\}$ states (five datasets per $M$); training runs for $T_\text{train}=15{,}000$ steps.

\emph{Task 2: Pusher}. We control a 7-DoF arm to push a cylinder to a target (23-dimensional state space, 7-dimensional action space). The dense shaping and randomized targets stress novelty detection in higher-dimensional state spaces; episodes are 100 steps. Offline datasets use $M\!\in\!\{1\text{k},2\text{k},5\text{k},10\text{k},20\text{k}\}$ (five per $M$); training runs for $T_\text{train}=2{,}000$ steps.

\emph{Task 3: Hopper}. We control a planar one-legged robot
that must hop forward without falling (11-dimensional state
space, 3-dimensional action space), emphasizing long-horizon
stability. The expert typically
makes a few hops before falling, so trajectories mix successful
and failing behavior; this makes state-space novelty less
informative for CRSAIL, while action-based gates can still
exploit the small action space. We therefore view Hopper as a
challenging, near worst-case benchmark for our method. Offline datasets use $M\!\in\!\{1\text{k},2\text{k},5\text{k},10\text{k},20\text{k}\}$ states
(five datasets per $M$); training runs for $T_\text{train}=10{,}000$ steps.
\begin{table}
  \centering
  \caption{Hyperparameters per method and environment. DAgger has no tunable hyperparameters. Abbreviations: \#M=number of models; $\tau_{\mathrm{agree}}$=agreement threshold; $\tau_{\mathrm{doubt}}$=doubt threshold; TQR=target query rate.}
  \label{tab:hyperparams_env_keyval}
  \setlength{\tabcolsep}{4pt}
  \begin{tabular}{l ll ll ll}
    \toprule
    & \multicolumn{2}{c}{\textbf{InvDP}} & \multicolumn{2}{c}{\textbf{Pusher}} & \multicolumn{2}{c}{\textbf{Hopper}} \\
    \cmidrule(lr){2-3}\cmidrule(lr){4-5}\cmidrule(lr){6-7}
    \textbf{Method} & \textbf{Key} & \textbf{Value} & \textbf{Key} & \textbf{Value} & \textbf{Key} & \textbf{Value} \\
    \midrule

    \multirow{3}{*}{EnsembleDAgger}
      & \#M & 5 & \#M & 5 & \#M & 5 \\
      & $\tau_{\mathrm{agree}}$ & 75\% & $\tau_{\mathrm{agree}}$ & 50\% & $\tau_{\mathrm{agree}}$ & 50\% \\
      & $\tau_{\mathrm{doubt}}$ & 0.01 & $\tau_{\mathrm{doubt}}$ & 0.03 & $\tau_{\mathrm{doubt}}$ & 0.1 \\
    \midrule
    \multirow{2}{*}{ThriftyDAgger}
      & \#M & 5 & \#M & 5 & \#M & 5 \\
      & TQR & 10\% & TQR & 40\% & TQR & 25\% \\
    \midrule
    \multirow{2}{*}{\textbf{CRSAIL (ours)}}
      & $K$ & 5 & $K$ & 5 & $K$ & 5 \\
      & $\alpha$ & 0.93 & $\alpha$ & 0.93 & $\alpha$ & 0.95 \\
    \bottomrule
  \end{tabular}
\end{table}

\begin{table}
  \centering
  \caption{Convergence rate, queries required to reach expert-level performance, and total queries made during 2,000 training steps on Pusher using CRSAIL with different $\alpha$ values ($K=5$).}
  \label{tab:alpha_sweep}
  \setlength{\tabcolsep}{3pt}
  \renewcommand{\arraystretch}{0.95}
  \small
  \resizebox{\columnwidth}{!}{
  \begin{tabular}{llrrrrr}
    \toprule
    \textbf{$\alpha$} & \textbf{Metric} & \textbf{1{,}000} & \textbf{2{,}000} & \textbf{5{,}000} & \textbf{10{,}000} & \textbf{20{,}000} \\
    \midrule
    \multirow{3}{*}{0.50}
      & Conv. (\%)   & 60 & 60 & 60 & 60 & 80 \\
      & Q$\to$Exp.   & 182$\pm$63 & 111$\pm$31 & 154$\pm$91 & 177$\pm$85 & \textbf{137$\pm$106} \\
      & Total Q.     & 226$\pm$186 & \textbf{117$\pm$65} & \textbf{193$\pm$87} & \textbf{161$\pm$90} & \textbf{139$\pm$97} \\
    \midrule
    \multirow{3}{*}{0.60}
      & Conv. (\%)   & 40 & 40 & 60 & 80 & 60 \\
      & Q$\to$Exp.   & \textbf{153$\pm$22} & \textbf{105$\pm$22} & \textbf{142$\pm$57} & 178$\pm$47 & 158$\pm$53 \\
      & Total Q.     & \textbf{205$\pm$126} & 164$\pm$90 & 220$\pm$70 & 222$\pm$119 & 194$\pm$138 \\
    \midrule
    \multirow{3}{*}{0.70}
      & Conv. (\%)   & 80 & 100 & 100 & 100 & 100 \\
      & Q$\to$Exp.   & 208$\pm$80 & 122$\pm$54 & 146$\pm$68 & \textbf{163$\pm$69} & 155$\pm$48 \\
      & Total Q.     & 451$\pm$422 & 253$\pm$86 & 260$\pm$179 & 366$\pm$252 & 213$\pm$79 \\
    \midrule
    \multirow{3}{*}{0.80}
      & Conv. (\%)   & 40 & 80 & 100 & 60 & 100 \\
      & Q$\to$Exp.   & 269$\pm$120 & 229$\pm$74 & 271$\pm$67 & 189$\pm$61 & 191$\pm$26 \\
      & Total Q.     & 242$\pm$112 & 238$\pm$79 & 643$\pm$304 & 186$\pm$130 & 254$\pm$28 \\
    \midrule
    \multirow{3}{*}{0.90}
      & Conv. (\%)   & 80 & 100 & 100 & 100 & 100 \\
      & Q$\to$Exp.   & 161$\pm$42 & 329$\pm$123 & 213$\pm$60 & 176$\pm$53 & 256$\pm$41 \\
      & Total Q.     & 289$\pm$108 & 657$\pm$242 & 448$\pm$132 & 336$\pm$94 & 549$\pm$205 \\
    \midrule
    \multirow{3}{*}{0.91}
      & Conv. (\%)   & 100 & 100 & 100 & 100 & 100 \\
      & Q$\to$Exp.   & 326$\pm$80 & 167$\pm$33 & 249$\pm$101 & 219$\pm$134 & 238$\pm$107 \\
      & Total Q.     & 800$\pm$276 & 327$\pm$125 & 599$\pm$211 & 544$\pm$299 & 602$\pm$326 \\
    \midrule
     \multirow{3}{*}{0.93}
      & Conv. (\%)   & 100 & 100 & 100 & 100 & 100 \\
      & Q$\to$Exp.   & 260$\pm$153 & 222$\pm$57 & 182$\pm$46 & 184$\pm$67 & 246$\pm$86 \\
      & Total Q.     & 708$\pm$422 & 761$\pm$367 & 463$\pm$258 & 398$\pm$146 & 495$\pm$155 \\
    \midrule
    \multirow{3}{*}{0.95}
      & Conv. (\%)   & 100 & 100 & 100 & 100 & 100 \\
      & Q$\to$Exp.   & 346$\pm$88 & 226$\pm$60 & 310$\pm$91 & 301$\pm$110 & 201$\pm$46 \\
      & Total Q.     & 794$\pm$245 & 748$\pm$521 & 882$\pm$396 & 539$\pm$160 & 492$\pm$176 \\
    \midrule
    \multirow{3}{*}{0.99}
      & Conv. (\%)   & 100 & 100 & 100 & 100 & 100 \\
      & Q$\to$Exp.   & 440$\pm$105 & 295$\pm$148 & 294$\pm$65 & 338$\pm$69 & 277$\pm$74 \\
      & Total Q.     & 1334$\pm$175 & 1226$\pm$139 & 1226$\pm$260 & 1085$\pm$215 & 1242$\pm$223 \\
    \bottomrule
  \end{tabular}}
\end{table}
\subsection{Effect of $\alpha$}
On Pusher, Table~\ref{tab:alpha_sweep} shows that the convergence probability increases with both \(\alpha\) and the initial dataset size \(M\); for \(\alpha \ge 0.9\) all runs converge for all \(M\), indicating that CRSAIL is robust to the choice of \(\alpha\). As expected from the calibration rule, larger $\alpha$ yields more total queries by lowering the target coverage level and marking more states as novel, while the dependence of queries-to-converge on $\alpha$ is much weaker, so higher $\alpha$ mainly drives additional queries after convergence. This robustness to \(\alpha\) is a practical advantage over existing AIL methods, which typically require careful per-task tuning of query thresholds. In practice, choosing a mid-range \(\alpha\) (e.g., \(0.9\)–\(0.95\)) balances convergence and budget.
Similar trends exist across the other tasks and more $\alpha$ values, but have been omitted due to space limitations in order to focus on algorithm comparisons.
\subsection{Effect of $K$}
On Pusher, we fix the miscoverage level at $\alpha = 0.93$ and sweep the
neighbor order $K$ (\cref{tab:K_sweep}). \emph{For every $K$ and initial dataset
size $M$, CRSAIL converges to expert-level reward on all runs}, and both the
queries-to-expert and total queries are nearly flat in $K$, indicating that
performance is largely insensitive to the neighbor order. Intuitively,
increasing $K$ averages over more neighbors and should mitigate the effect of
isolated outliers at the cost of slightly more queries; the empirical flatness
suggests that such isolated outliers are rare in these sequential trajectories
and that even a small neighborhood already captures the relevant local geometry
of the expert data. Similar behavior is observed on the other environments and
omitted for space.

\begin{table}
  \centering
  \caption{Queries required for CRSAIL to reach expert-level performance and total queries made during 2,000 training steps on Pusher, for varying initial expert dataset sizes, with a sweep over $K$ ($\alpha=0.93$).}
  \label{tab:K_sweep}
  \setlength{\tabcolsep}{3pt}
  \renewcommand{\arraystretch}{0.95}
  \small
  \resizebox{\columnwidth}{!}{
  \begin{tabular}{llrrrrr}
    \toprule
    \textbf{$K$} & \textbf{Metric} & \textbf{1{,}000} & \textbf{2{,}000} & \textbf{5{,}000} & \textbf{10{,}000} & \textbf{20{,}000} \\
    \midrule
    \multirow{2}{*}{1}
      & Q$\to$Exp.   & 238$\pm$115 & 222$\pm$63 & 228$\pm$66 & 211$\pm$47 & \textbf{245$\pm$52} \\
      & Total Q.     & 600$\pm$155 & 583$\pm$130 & 510$\pm$151 & 383$\pm$138 & \textbf{459$\pm$177} \\
    \midrule
    \multirow{2}{*}{3}
      & Q$\to$Exp.   & 265$\pm$50 & 193$\pm$74 & 329$\pm$104 & 260$\pm$83 & 283$\pm$167 \\
      & Total Q.     & 496$\pm$197 & 536$\pm$180 & 875$\pm$438 & 429$\pm$271 & 552$\pm$393 \\
    \midrule
    \multirow{2}{*}{5}
      & Q$\to$Exp.   & 260$\pm$153 & 222$\pm$57 & \textbf{182$\pm$46} & 184$\pm$67 & 246$\pm$86 \\
      & Total Q.     & 708$\pm$422 & 761$\pm$367 & \textbf{463$\pm$258} & 398$\pm$146 & 495$\pm$155 \\
    \midrule
    \multirow{2}{*}{7}
      & Q$\to$Exp.   & \textbf{189$\pm$75} & \textbf{176$\pm$19} & 285$\pm$120 & 208$\pm$74 & 306$\pm$131 \\
      & Total Q.     & \textbf{427$\pm$93} & \textbf{530$\pm$156} & 537$\pm$248 & 487$\pm$143 & 699$\pm$369 \\
    \midrule
    \multirow{2}{*}{9}
      & Q$\to$Exp.   & 265$\pm$56 & 212$\pm$42 & 210$\pm$97 & \textbf{147$\pm$58} & 348$\pm$87 \\
      & Total Q.     & 684$\pm$345 & 567$\pm$360 & 532$\pm$270 & \textbf{357$\pm$133} & 516$\pm$156 \\
    \bottomrule
  \end{tabular}}
\end{table}
\subsection{Comparisons}
As previously mentioned, we compare our work against DAgger, EnsembleDAgger and ThriftyDAgger through our suite of environments.
 
\begin{table}[!ht]
  \centering
  \caption{Convergence rate, queries required to reach expert-level performance, and total queries across tasks and initial expert dataset sizes. We abbreviate EnsembleDAgger as \textbf{Ensemble} and ThriftyDAgger as \textbf{Thrifty}. $K=5$ for all environments; $\alpha_{\mathrm{InvDP}}=\alpha_{\mathrm{Pusher}}=0.93;\ \alpha_{\mathrm{Hopper}}=0.95$}
  \label{tab:merged_all_tasks}
  \renewcommand{\arraystretch}{0.95}
  \small
\begin{subtable}{\linewidth}
\captionsetup{aboveskip=0.5\baselineskip, belowskip=0.5\baselineskip}
    \centering
    \caption{Inverted Double Pendulum ($T_{\mathrm{train}}=15{,}000$ steps)}
    \label{tab:invdp_main}
    \begingroup
    \setlength{\tabcolsep}{3pt}
    \resizebox{\linewidth}{!}{
    \begin{tabular}{llrrrrr}
      \toprule
      \textbf{Method} & \textbf{Metric} & \textbf{1{,}000} & \textbf{2{,}000} & \textbf{3{,}000} & \textbf{5{,}000} & \textbf{10{,}000} \\
      \midrule
      \multirow{3}{*}{DAgger}
        & Conv. (\%)   & 100 & 100 & 100 & 100 & 100 \\
        & Q$\to$Exp.   & 1903$\pm$1379 & 2324$\pm$1211 & 3433$\pm$2497 & 3306$\pm$2662 & 642$\pm$878 \\
        & Total Q.     & 15545$\pm$119 & 15606$\pm$251 & 15650$\pm$151 & 15461$\pm$143 & 15508$\pm$68 \\
      \midrule
      \multirow{3}{*}{Ensemble}
        & Conv. (\%)   & 100 & 80 & 100 & 80 & 80 \\
        & Q$\to$Exp.   & 1161$\pm$874 & 583$\pm$504 & 910$\pm$735 & 815$\pm$612 & 512$\pm$109 \\
        & Total Q.     & 1960$\pm$1425 & 1537$\pm$954 & 1400$\pm$740 & 1571$\pm$744 & 1318$\pm$717 \\
      \midrule
      \multirow{3}{*}{Thrifty}
        & Conv. (\%)   & 100 & 100 & 80 & 80 & 80 \\
        & Q$\to$Exp.   & 175$\pm$77 & 303$\pm$211 & 246$\pm$58 & 813$\pm$988 & \textbf{282$\pm$155} \\
        & Total Q.     & 1843$\pm$678 & 856$\pm$323 & 1975$\pm$656 & 1342$\pm$748 & 1581$\pm$781 \\
      \midrule
      \multirow{3}{*}{\textbf{CRSAIL}}
        & Conv. (\%)   & 100 & 100 & 100 & 100 & 100 \\
        & Q$\to$Exp.   & \textbf{164$\pm$42} & \textbf{206$\pm$30} & \textbf{190$\pm$30} & \textbf{233$\pm$45} & 466$\pm$233 \\
        & Total Q.     & \textbf{254$\pm$69} & \textbf{338$\pm$73} & \textbf{484$\pm$247} & \textbf{866$\pm$355} & \textbf{784$\pm$172} \\
      \bottomrule
    \end{tabular}}
    \endgroup
\end{subtable}
\begin{subtable}{\linewidth}
\captionsetup{aboveskip=0.5\baselineskip, belowskip=0.5\baselineskip}
    \centering
    \caption{Pusher ($T_{\mathrm{train}}=2{,}000$ steps)}
    \label{tab:pusher_main}
    \begingroup
    \setlength{\tabcolsep}{3.5pt}
    \resizebox{\linewidth}{!}{
    \begin{tabular}{llrrrrr}
      \toprule
      \textbf{Method} & \textbf{Metric} & \textbf{1{,}000} & \textbf{2{,}000} & \textbf{5{,}000} & \textbf{10{,}000} & \textbf{20{,}000} \\
      \midrule
      \multirow{3}{*}{DAgger}
        & Conv. (\%)   & 100 & 100 & 100 & 100 & 100 \\
        & Q$\to$Exp.   & 480$\pm$172 & 340$\pm$75 & 340$\pm$102 & 700$\pm$141 & 580$\pm$240 \\
        & Total Q.     & 2000$\pm$0 & 2000$\pm$0 & 2000$\pm$0 & 2000$\pm$0 & 2000$\pm$0 \\
      \midrule
      \multirow{3}{*}{Ensemble}
        & Conv. (\%)   & 100 & 100 & 100 & 100 & 100 \\
        & Q$\to$Exp.   & 293$\pm$203 & 274$\pm$154 & 237$\pm$126 & 217$\pm$74 & 469$\pm$109 \\
        & Total Q.     & 1415$\pm$221 & 1147$\pm$82 & 911$\pm$130 & 957$\pm$45 & 987$\pm$39 \\
      \midrule
      \multirow{3}{*}{Thrifty}
        & Conv. (\%)   & 100 & 100 & 80 & 40 & 20 \\
        & Q$\to$Exp.   & 490$\pm$87 & 651$\pm$154 & 854$\pm$162 & 1098$\pm$62 & 1327$\pm$0 \\
        & Total Q.     & 1294$\pm$56 & 1308$\pm$30 & 1283$\pm$81 & 1280$\pm$48 & 1290$\pm$33 \\
      \midrule
      \multirow{3}{*}{\textbf{CRSAIL}}
        & Conv. (\%)   & 100 & 100 & 100 & 100 & 100 \\
        & Q$\to$Exp.   & \textbf{260$\pm$153} & \textbf{222$\pm$57} & \textbf{182$\pm$46} & \textbf{184$\pm$67} & \textbf{246$\pm$86} \\
        & Total Q.     & \textbf{708$\pm$422} & \textbf{761$\pm$367} & \textbf{463$\pm$258} & \textbf{398$\pm$146} & \textbf{495$\pm$155} \\
      \bottomrule
    \end{tabular}}
    \endgroup
\end{subtable}
\begin{subtable}{\linewidth}
\captionsetup{aboveskip=0.5\baselineskip, belowskip=0.5\baselineskip}
    \centering
    \caption{Hopper ($T_{\mathrm{train}}=10{,}000$ steps)}
    \label{tab:hopper_main}
    \begingroup
    \setlength{\tabcolsep}{3.5pt}
    \resizebox{\linewidth}{!}{
    \begin{tabular}{llrrrrr}
      \toprule
      \textbf{Method} & \textbf{Metric} & \textbf{1{,}000} & \textbf{2{,}000} & \textbf{5{,}000} & \textbf{10{,}000} & \textbf{20{,}000} \\
      \midrule
      \multirow{3}{*}{DAgger}
        & Conv. (\%)   & 100 & 100 & 100 & 100 & 100 \\
        & Q$\to$Exp.   & 3930$\pm$619 & 3728$\pm$601 & 3040$\pm$870 & 3294$\pm$696 & 4576$\pm$657 \\
        & Total Q.     & 10198$\pm$95 & 10242$\pm$247 & 10249$\pm$170 & 10091$\pm$59 & 10192$\pm$135 \\
      \midrule
      \multirow{3}{*}{Ensemble}
        & Conv. (\%)   & 100 & 100 & 100 & 100 & 100 \\
        & Q$\to$Exp.   & 1969$\pm$178 & 1839$\pm$318 & 2043$\pm$324 & 1990$\pm$446 & 2804$\pm$320 \\
        & Total Q.     & 8879$\pm$193 & 8722$\pm$217 & 8966$\pm$155 & 8806$\pm$315 & 8576$\pm$137 \\
      \midrule
      \multirow{3}{*}{Thrifty}
        & Conv. (\%)   & 100 & 100 & 80 & 100 & 60 \\
        & Q$\to$Exp.   & \textbf{1514$\pm$487} & \textbf{1100$\pm$403} & \textbf{1537$\pm$247} & \textbf{1566$\pm$134} & \textbf{1806$\pm$487} \\
        & Total Q.     & 2991$\pm$118 & 2873$\pm$183 & \textbf{2827$\pm$164} & \textbf{2740$\pm$182} & \textbf{2650$\pm$43} \\
      \midrule
      \multirow{3}{*}{\textbf{CRSAIL}}
        & Conv. (\%)   & 100 & 80 & 100 & 100 & 100 \\
        & Q$\to$Exp.   & 1821$\pm$103 & 2299$\pm$300 & 2355$\pm$279 & 2694$\pm$408 & 2519$\pm$628 \\
        & Total Q.     & \textbf{2928$\pm$298} & \textbf{2578$\pm$296} & 4038$\pm$978 & 5760$\pm$1286 & 3388$\pm$702 \\
      \bottomrule
    \end{tabular}}
    \endgroup
\end{subtable}
\end{table}

\Cref{tab:merged_all_tasks} compares convergence, queries-to-expert, and total expert queries across tasks and initial expert set sizes. On \emph{Inverted Double Pendulum} and \emph{Pusher}, \textsc{CRSAIL} achieves \textbf{100\% convergence for every \(M\)} while querying dramatically less than all baselines. On \emph{Pusher} specifically, \textsc{CRSAIL} is uniformly the most query-efficient method \emph{for all \(M\)}—it requires the fewest queries to reach expert-level performance and the fewest total queries. On \emph{InvDP}, it uses the fewest \emph{total} queries at every \(M\) and attains the fewest \emph{queries to converge to expert} in \(4/5\) cases (second-best in the remaining setting). By contrast, ThriftyDAgger struggles to converge on \emph{Pusher} as \(M\) increases (convergence falling to \(80\%,40\%,20\%\)) despite a relatively high target query rate \(\mathrm{TQR}=0.4\); increasing \(\mathrm{TQR}\) could improve convergence but would inflate query counts toward DAgger-like levels.

\noindent On \emph{Hopper}—our most challenging domain—\textsc{CRSAIL} remains modestly successful: it converges in \(\mathbf{24/25}\) runs overall and achieves the lowest \emph{total} query counts at the two smallest initial datasets (\(M{=}1\mathrm{k},2\mathrm{k}\)), while remaining competitive at larger \(M\). Unless noted otherwise, \(K{=}5\) and the threshold parameters \(\alpha\) follow the table caption. We report \emph{queries until convergence to expert} as the mean over only those runs that did converge (non-converged runs are excluded from that average).

\begin{table}[!ht]
  \centering
  \caption{Mean peak reward and query efficiency across environments.
  Values are mean $\pm$ std.\ across all dataset sizes (25 total seeds). Best reward is the mean of the highest reward earned during each training run.}
  \label{tab:env_method_merged}
  \setlength{\tabcolsep}{4pt}
  \renewcommand{\arraystretch}{0.96}
  \resizebox{\columnwidth}{!}{
  \begin{tabular}{l|l r r r}
    \toprule
    \textbf{Task} & \textbf{Method} &
    \shortstack{\textbf{Best Reward}\\\textbf{(\% Expert)}} &
    \shortstack{\textbf{Total Queries}\\\textbf{(\% DAgger)}} &
    \shortstack{\textbf{Total Queries}\\\textbf{(\% SOTA)}} \\
    \midrule
    \multirow{4}{*}{InvDP}
    & DAgger   & \textbf{108.9\% $\pm$ 1.2\%} & 100.0\% $\pm$ 1.4\% & 1023.9\% $\pm$ 444.2\% \\
      & Ensemble & 104.5\% $\pm$ 7.9\% & 10.0\% $\pm$ 6.1\% & 102.5\% $\pm$ 77.0\% \\
      & Thrifty  & 104.2\% $\pm$ 8.6\% & 9.8\% $\pm$ 4.2\%  & 100.0\% $\pm$ 0.0\% \\
      & \textbf{CRSAIL} & 107.6\% $\pm$ 7.0\% & \textbf{3.4\% $\pm$ 1.0\%} & \textbf{34.7\% $\pm$ 18.1\%} \\
      
    \midrule
    \multirow{4}{*}{Pusher}
    & DAgger   & 103.8\% $\pm$ 1.1\%  & 100.0\% $\pm$ 0.0\% & 184.6\% $\pm$ 21.0\% \\
      & Ensemble & \textbf{104.5\% $\pm$ 1.2\%} & 54.2\% $\pm$ 6.2\% & 100.0\% $\pm$ 0.0\% \\
      & Thrifty  & 97.4\% $\pm$ 3.6\%  & 64.5\% $\pm$ 2.6\% & 119.1\% $\pm$ 14.4\% \\
      & \textbf{CRSAIL} & 103.5\% $\pm$ 1.7\% & \textbf{28.3\% $\pm$ 14.6\%} & \textbf{52.2\% $\pm$ 27.6\%} \\
      \midrule
      \multirow{4}{*}{Hopper}
        & DAgger   & 107.2\% ± 29.8\% & 100.0\% ± 2.2\% & 361.9\% ± 21.3\% \\ 
        & Ensemble & \textbf{112.1\% ± 24.1\%} & 86.2\% ± 2.5\% & 312.1\% ± 19.3\% \\ 
        & Thrifty  & 98.4\% ± 23.4\%  & \textbf{27.6\% ± 1.5\%} & \textbf{100.0\% ± 0.0\%} \\ 
        & CRSAIL   & 102.2\% ± 24.6\% & 36.7\% ± 8.0\% & 132.9\% ± 29.8\% \\  
    \bottomrule
  \end{tabular}}
\end{table}
\Cref{tab:env_method_merged} aggregates across all $M$ values to paint a clear picture: In Inverted Double Pendulum and Pusher, \emph{CRSAIL requests $66.3\%$ and $47.8\%$ fewer expert labels respectively compared to the previous state of the art}, without a meaningful loss of reward. In the hopper task, which as discussed previously is the most challenging to our method, CRSAIL still holds its own: We require fewer queries than EnsembleDAgger, and we achieve better performance than ThriftyDAgger.
Note that state-of-the-art is chosen as the algorithm (aside from CRSAIL) that makes the fewest queries on average.
\begin{figure}
  \centering
  \begin{subfigure}{0.49\linewidth}
    \includegraphics[width=\linewidth]{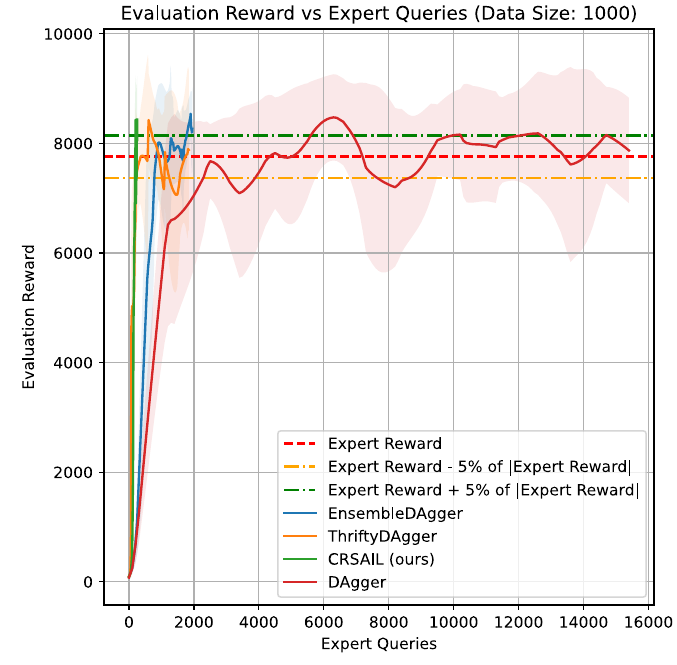}
\caption{}\label{fig:reward_vs_queries}
  \end{subfigure}
  \begin{subfigure}{0.49\linewidth}
  
    \includegraphics[width=\linewidth]{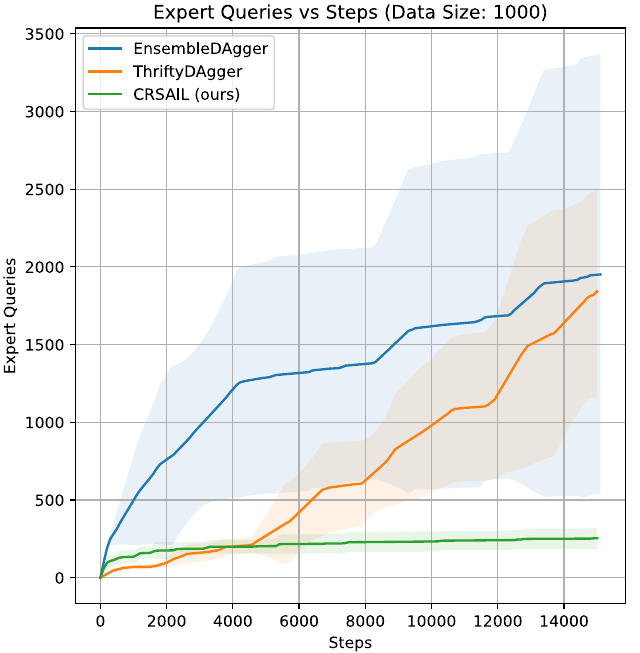}
    \caption{}\label{fig:queries_vs_steps}
  \end{subfigure}
  \caption{Querying metrics throughout training, aggregated between all initial datasets of the same size (showing $M=1000$ as a representative example) for InvDP. (a) shows the obtainable reward with regard to the number of queries, and (b) shows the number of queries made during the training process. Filled areas show standard deviation.}
  \label{fig:metrics-horizontal}
\end{figure}

On Inverted Double Pendulum, \cref{fig:metrics-horizontal}\subref{fig:reward_vs_queries} plots how
fast each method approaches expert-like behavior as a function of the
cumulative number of expert queries. Because all methods share the same step
budget \(T_{\mathrm{train}}\), a vertical slice at a given query budget \(B_0\)
can be interpreted as a fixed-budget comparison: it shows the reward each
method achieves before exceeding \(B_0\). As we can see, CRSAIL converges to
the expert the fastest and would score higher than other algorithms for any
fixed query budget in this range. We can also see that CRSAIL's curve terminates
earliest, indicating that the method stops querying once the desired behavior
is achieved. \Cref{fig:metrics-horizontal}\subref{fig:queries_vs_steps} shows
the number of queries across
training steps. In contrast to other algorithms, CRSAIL intuitively starts with a high query rate as additional information is needed and slowly plateaus. Qualitatively similar trends are observed on Pusher
and Hopper, but we omit those plots for space. DAgger is omitted as it coincides with the line \(y=x\), which dominates the vertical scale.
\begin{figure}[!t]
    \centering
    \includegraphics[width=0.66\linewidth]{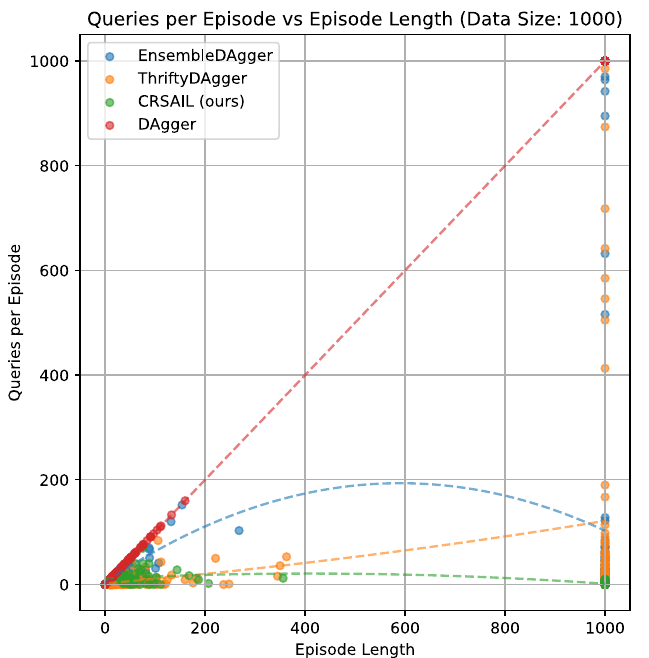}
    \caption{Number of queries made in each episode during training, based on the length of the episode for the inverted double pendulum task.}
  \label{fig:query_vs_length}
\end{figure}
On InvDP, the initial expert dataset mostly consists of successful episodes, and going off-policy results in failure. A query-efficient method should need fewer queries in longer episodes whose states are already well represented; \cref{fig:query_vs_length} confirms this effect. The effect is stronger for larger initial datasets, which contain more successful trajectories.

\section{Conclusions and Future Work}
We presented CRSAIL, a conformal prediction–based framework for query-efficient active imitation learning. Using distance-based nonconformity scores and a principled calibration step, CRSAIL reduces redundant expert queries while maintaining robust convergence across tasks. Our experiments on Inverted Double Pendulum and Pusher demonstrate that CRSAIL is significantly more query efficient than state-of-the-art baselines, while maintaining expert-like performance. Moreover, ablations confirm that CRSAIL is robust to hyperparameter choices and adapts naturally to different dataset sizes and episode structures.

Future work includes incorporating action similarity into the distance metric to better handle critical regions where the policy is less stable, exploring
time-varying miscoverage along recalibration to obtain
principled control over the decay of the query rate, and applying conformal calibration to other out-of-distribution scores.

\bibliographystyle{IEEEtran}   
\bibliography{refs}   

\end{document}